\documentclass[11pt,a4paper]{article}

\usepackage[T1]{fontenc}
\usepackage[utf8]{inputenc}
\usepackage{lmodern}
\usepackage{microtype}
\usepackage[margin=1in]{geometry}
\usepackage{graphicx}
\usepackage{float}
\usepackage{booktabs}
\usepackage{tabularx}
\usepackage{multirow}
\usepackage{amsmath}
\usepackage{xcolor}
\usepackage{listings}
\usepackage[numbers,sort&compress]{natbib}
\usepackage[hidelinks]{hyperref}
\usepackage[nameinlink,capitalise,noabbrev]{cleveref}

\graphicspath{{figures/}}

\usepackage{hyperref}
\hypersetup{colorlinks=true, citecolor=blue}
\usepackage{array}
\usepackage{multirow}
\usepackage{multicol}
\usepackage{booktabs}
\usepackage{tabularx}
\usepackage{threeparttable}
\usepackage{hhline}
\usepackage{colortbl}
\usepackage{arydshln}

\usepackage{algorithm}
\usepackage{algorithmic}
\usepackage{pifont}

\usepackage[T1]{fontenc}
\usepackage{textcomp}
\usepackage{textgreek}
\usepackage{csquotes}
\usepackage{balance}

\usepackage{graphicx}
\DeclareGraphicsExtensions{.pdf,.jpeg,.jpg,.png,.eps}
\usepackage{wrapfig}
\usepackage{subcaption}
\usepackage{float}
\usepackage{adjustbox}
\usepackage{rotating}

\title{The Punjabi Geet Toxicity Dataset (PGT-FG): Fine-Grained Multimodal Annotations for Toxic Content in Punjabi Songs}

\author{
  Lakhjeet Kaur\thanks{First author. Department of Computer Science, Guru Gobind Singh Indraprastha University, Delhi, India. Email: \texttt{jeetuppal9@gmail.com}} \\
  \and
    Jaspreet Kaur\thanks{Department of Computer Science, Malwa College Bathinda, Punjab, India. Email: \texttt{jaspreetbehniwal8@gmail.com}} \\
  \and
  Komal Rani\thanks{Department of Management, Akal University Talwandi Sabo, Punjab, India. Email: \texttt{komalkamboj7861@gmail.com}} \\
  \and
  Yougal Attri\thanks{Independent Researcher, New Delhi, India. Email: \texttt{yougalattri17@gmail.com}} \\
    \and
   Ramakant Kumar\thanks{Department of Computer Science, Galgotias College of Engineering \& Technology, India. Email: \texttt{ramakant.kumar@galgotiacollege.edu}}\\
  \and
   Jaydeep Kishore\thanks{Department of Computer Science, UPES Dehradun, India. Email: \texttt{jaydeep.kishore@ddn.upes.ac.in}} \\
  \and
  Pravin Kumar\thanks{Corresponding author. Department of Computer Science, UPES Dehradun, India. Email: \texttt{pravin.kumar@ddn.upes.ac.in}}
  \date{}
}

\begin{document}

\maketitle
\begin{abstract}

\textbf{Background:}
Resources for toxicity detection in Punjabi remain limited, particularly for
musical content that combines linguistic and multimodal information. We
present the Punjabi Geet Toxicity Dataset
(PGT-FG), a segment-level resource developed
to support research on toxicity detection in Punjabi musical content.

\textbf{Methods:}
The dataset was constructed from publicly available Punjabi songs collected
from YouTube. An initial collection of 620 songs resulted in 465 successfully
downloaded songs and 3,402 preliminary segments from 450 songs. After
quality control and annotation, the final dataset contains 2,396 unique
segments from 397 songs. The original annotation process generated 2,870
annotation records from four Punjabi-native speaker annotators, including a
shared subset of 158 segments independently annotated by all four annotators.
Each annotation includes binary toxicity, eight fine-grained toxicity
categories, severity, sentiment, mode, and an uncertainty flag. Fleiss'
Kappa was calculated on the shared subset to assess inter-annotator agreement,
and majority-based resolution was used to obtain final segment-level labels.

\textbf{Results:}
Among the 2,396 unique segments, 705 (29.42\%)
were labelled toxic and 1,691 (70.58\%) non-toxic. Violence was the
most frequent toxicity category, occurring in 539 toxic segments
(76.45\%), followed by gang references in 227 (32.20\%) and drug
references in 133 (18.87\%). The overall toxicity label achieved
substantial inter-annotator agreement ($\kappa = 0.6521$).

\textbf{Conclusions:}
The Punjabi Geet Toxicity Dataset provides a fine-grained resource for Punjabi
toxicity detection and multimodal content analysis. The resolved dataset
contains one final annotation record for each of the 2,396 unique segments
and is publicly available through Zenodo at
\url{https://doi.org/10.5281/zenodo.23040156}. Original lyric text, lyric
excerpts, and copyrighted audio or video content are not redistributed.

\end{abstract}


\section{Introduction}

Music is more than a form of entertainment; it is also a way of expressing emotions, identities, experiences, and social perspectives. Song lyrics can reflect everyday life and culture, but they can also contain language associated with violence, misogyny, drug use, gang culture, profanity, and other forms of harmful expression. The automatic identification of such content has become an important research problem in Natural Language Processing (NLP), particularly with the growth of online platforms and user-generated content. Previous research has explored the automatic detection of hate speech, abusive language, and other harmful forms of communication, showing the importance of computational methods for analysing large amounts of textual data \cite{khurana-etal-2022-animojity,kapil-ekbal-2020-multidomain}. However, detecting harmful content in music lyrics presents additional challenges. Lyrics often rely on context, metaphors, slang, repetition, storytelling, and culturally specific expressions. Therefore, datasets developed for general online communication may not adequately represent the characteristics of music lyrics. The quality and reliability of the underlying dataset are also important, as differences in annotation schemes and data collection practices can substantially affect toxicity detection systems \cite{wiegand-etal-2019-biased}.

Punjabi music provides an important setting for studying this problem. Punjabi popular music has developed a strong presence both within South Asia and among Punjabi communities around the world. Research on Punjabi popular music has highlighted its role in shaping cultural identity and connecting ideas of Punjab with audiences in different parts of the world \cite{schreffler2012migration}. Punjabi music also continues to play an important role in representing regional identity, social experiences, aspirations, and changing cultural practices \cite{szivak2022changing}. At the same time, some areas of contemporary Punjabi music have received attention because of themes related to violence, aggression, drugs, weapons, and gangster culture. Recent studies have examined the representation and interpretation of violent and aggressive themes in Punjabi popular songs \cite{singh2025violent,vanderbeek2025sonic}. These observations should not be taken to mean that Punjabi music as a whole is toxic. Instead, they demonstrate the need for systematic analysis at the level of individual songs and lyrical segments. Such analysis is particularly challenging because Punjabi lyrics may include colloquial expressions, transliterated text, code-mixing, regional vocabulary, and culturally specific meanings.

Although research on toxic and abusive language has produced a growing number of datasets, most existing resources focus on online communication such as social media posts, comments, and discussion forums \cite{song2021large,risch2021data}. Fine-grained annotation has also been recognized as valuable for distinguishing between different forms of harmful language rather than treating all harmful content as a single category \cite{kurrek2020towards}. However, music lyrics remain comparatively underrepresented in this area, and Punjabi music presents an additional gap. Existing Punjabi language resources support several NLP tasks, but there is limited availability of a dedicated dataset focused specifically on toxicity in Punjabi songs. To the best of our knowledge, there is no publicly available resource that
combines Punjabi song lyrics, segment-level multi-category toxicity
annotations, and corresponding audio-visual source information for this task. This gap makes it difficult to train, evaluate, and compare models specifically designed for Punjabi music toxicity detection. A dedicated dataset can therefore provide a useful foundation for research at the intersection of Punjabi NLP, harmful-language detection, music analysis, and multimodal artificial intelligence.

To address this gap, we introduce \textit{the Punjabi Geet Toxicity Dataset
(PGT-FG)}, a multimodal dataset developed specifically for research on
toxicity detection in Punjabi songs. The dataset contains 397 unique songs
and 2,396 annotated lyric segments. During annotation and preprocessing,
each segment was associated with its corresponding lyric transcript,
audio-visual source material, and relevant song metadata. The public release
provides the segment identifiers, temporal information, source metadata,
and annotation labels, while copyrighted lyric, audio, and video content
are not redistributed. The corresponding audio and visual source material
can be accessed through the provided public source links, subject to their
continued availability and applicable copyright and platform terms. The
annotation scheme includes an overall toxicity label, sentiment labels,
and fine-grained categories representing different forms of potentially
harmful content. The song segments were divided among four annotators, with
each segment outside the shared subset assigned to a single annotator. A
shared subset of 158 segments was independently annotated by all four
annotators, allowing the dataset to capture different judgments about
potentially ambiguous lyrical content. By connecting lyric segments with
their corresponding audio and visual source information, the dataset can
support both text-based approaches and future multimodal approaches that
combine linguistic, acoustic, and visual information. The resource is
intended to provide a common foundation for developing and evaluating
computational methods for Punjabi music toxicity detection.

The main contributions of this work are as follows:

\begin{itemize}
    \item \textbf{A dedicated Punjabi music toxicity resource:}
    We introduce an annotated collection specifically designed for studying
    toxic content in Punjabi songs.

    \item \textbf{Fine-grained toxicity annotations:}
    The resource provides overall toxicity labels together with multiple
    categories of harmful content, enabling more detailed analysis than a
    simple toxic/non-toxic classification.

    \item \textbf{A multimodal resource:}
The collection connects annotated lyric segments with corresponding audio
and visual source information through publicly available source links,
supporting future research on multimodal toxicity detection without
redistributing the original copyrighted media.

    \item \textbf{A resource for future Punjabi NLP research:}
    The corpus provides a foundation for developing and evaluating machine
    learning, NLP, and multimodal models for Punjabi music analysis and
    toxicity detection.
\end{itemize}

\section{Related Datasets}
\label{sec:related_datasets}

Several datasets have been developed for toxicity, hate speech, and offensive
content detection across different languages and modalities. However, most
existing resources focus either on text-based social media content or on
multimodal content in languages such as English and Hindi. This section
reviews the datasets most closely related to the proposed PGT-FG and highlights the gap addressed by our work.

\subsection{Multimodal and Video-Based Datasets}

ToxCMM, introduced by Maity et al.~\cite{maity2024toxvidlm}, is the closest
existing work to our dataset. It is a multimodal dataset designed for toxicity
detection in Hindi--English code-mixed videos. The dataset contains 931
YouTube videos segmented into 4,021 utterance-level clips. Each utterance is
annotated for toxicity, sentiment, and offensive severity. The dataset
incorporates three modalities: text transcripts, audio, and visual content.
Although ToxCMM demonstrates the importance of multimodal toxicity detection
for Indian languages, it focuses on Hindi--English conversational videos
rather than Punjabi songs.

HateMM~\cite{das2023hatemm} is another multimodal dataset developed for hate
speech detection in videos. It consists of approximately 43 hours of
English-language videos collected from BitChute. The dataset provides
hate/non-hate labels and identifies portions of videos that contribute to the
classification. While it demonstrates the usefulness of multimodal
information for hate speech detection, it is restricted to English and does
not specifically address code-mixed or music-based content.

MultiHateClip~\cite{wang2024multihateclip} is a multilingual benchmark for
hateful video detection containing video clips from YouTube and Bilibili.
It focuses primarily on gender-related hateful content and includes English
and Chinese data. The dataset provides hate, offensive, and normal labels.
However, Punjabi is not represented, and the dataset does not specifically
target toxicity in musical content.

\subsection{Text and Multimodal Toxicity Datasets}

HateXplain~\cite{mathew2021hatexplain} is a widely used text-based hate speech
dataset containing 20,148 social media posts. Each post is annotated as
hate, offensive, or normal and additionally contains information about the
target community and word-level rationales. Although its explainable
annotations are valuable for toxicity research, the dataset is primarily
textual and does not contain audio or video information.

The Offensive Language Identification Dataset (OLID)~\cite{zampieri2019olid}
contains more than 14,000 English tweets and provides hierarchical
annotations for offensive language, target identification, and target type.
Similarly, HASOC~\cite{modha2019hasoc} provides multilingual hate speech and
offensive content annotations for English, Hindi, and German. These datasets
are important resources for textual toxicity detection, but they do not
provide the multimodal and music-specific information required for analysing
Punjabi songs.

The Hateful Memes dataset~\cite{kiela2020hateful} contains 10,000 memes
consisting of images with embedded text. It was developed to study multimodal
hate detection and demonstrates that combining visual and textual information
can be important for identifying hateful content. However, it is English
focused and does not include audio or video modalities.

\subsection{Audio and Punjabi Language Resources}

MuTox~\cite{costa2024mutox} is a large multilingual audio-based toxicity
dataset containing toxic and non-toxic spoken utterances across 30 languages.
It demonstrates the value of analysing toxicity directly from speech rather
than relying only on transcripts. Nevertheless, Punjabi is not included and
the dataset does not incorporate visual information or music-specific
content.

Punjabi-specific NLP resources are available across several language
understanding tasks, although resources focused specifically on toxicity
detection remain limited. ShahiEmotion
\cite{nawaz2026shahiemotion} provides a large Punjabi emotion dataset
containing 30,379 Shahmukhi sentences annotated with seven emotion
categories. Although this resource contributes to Punjabi language
understanding, it addresses emotion recognition rather than toxicity and is
text-only.

\subsection{Comparison with Existing Datasets}

Table~\ref{tab:dataset_comparison} summarizes the major differences between
existing datasets and PGT-FG.
\begin{table}[htbp]
\centering
\caption{Comparison of PGT-FG with existing toxicity-related datasets.}
\label{tab:dataset_comparison}
\resizebox{\columnwidth}{!}{%
\begin{tabular}{|l|c|c|c|c|}
\hline
\rowcolor{blue!15}
\textbf{Dataset} & \textbf{Language} & \textbf{Modalities} &
\textbf{Samples (Segments)} & \textbf{Label Types} \\
\hline
ToxCMM/ToxVidLM & Hindi--English & Text, Audio, Video & 4,021 &
Toxicity, Sentiment, Severity \\
\hline
HateMM & English & Text, Audio, Video & 43 hrs &
Hate/Non-hate \\
\hline
MultiHateClip & English, Chinese & Video & 2,000 &
Hate/Offensive/Normal \\
\hline
HateXplain & English & Text & 20,148 &
Hate/Offensive/Normal, Target, Rationales \\
\hline
OLID & English & Text & 14,000+ &
Offensive, Target, Target Type \\
\hline
HASOC & English, Hindi, German & Text & 17,657 &
Hate/Offensive/Non-offensive \\
\hline
Hateful Memes & English & Text, Image & 10,000 &
Hateful/Non-hateful \\
\hline
MuTox & Multilingual & Audio & 120,000+ &
Toxic/Non-toxic \\
\hline
ShahiEmotion & Punjabi & Text & 30,379 &
7 Emotion Classes \\
\hline
\textbf{PGT-FG (ours)} & \textbf{Punjabi--English} &
\textbf{Text, Audio, Video (via source links)} & \textbf{2,396} &
\textbf{Toxicity, Sentiment, Severity, Mode, Toxicity Categories} \\
\hline
\end{tabular}%
}
\end{table}

The comparison shows that existing datasets address important aspects of
toxicity, hate speech, offensive language, and multimodal content, but
there is a lack of resources specifically designed for Punjabi--English
musical content. ToxCMM is closely related to the proposed work because it
combines text, audio, and video with toxicity, sentiment, and severity
annotations. However, it focuses on Hindi--English code-mixed videos rather
than Punjabi songs. Similarly, MuTox provides multilingual audio-based
toxicity annotations but does not incorporate visual information or
music-specific content. Existing Punjabi resources such as ShahiEmotion
primarily focus on emotion recognition and are text-only.

PGT-FG addresses this gap by providing segment-level multimodal data from Punjabi songs, combining lyrics, audio,
and visual information. Each segment is annotated for overall toxicity,
sentiment, severity, mode, and fine-grained toxic-content categories,
including violence, misogyny, casteism, religious content, drugs, gang
references, profanity, and other toxic content. The inclusion of sentiment,
severity, and mode annotations alongside toxicity labels enables more
detailed analysis of toxic content in Punjabi musical discourse. The
combination of Punjabi--English lyrics, multimodal content, and
fine-grained toxicity annotations provides a resource specifically targeted
toward toxicity analysis in Punjabi musical content.

\section{Data Collection}
\label{sec:data_collection}

PGT-FG was constructed using Punjabi songs collected from
YouTube. YouTube was selected as the primary source because it provides a
large and diverse collection of Punjabi music videos containing lyrics,
audio, and visual information. The data collection process was designed to
obtain broad coverage of Punjabi musical content rather than selecting songs
from only a small number of artists or a single type of content.

\begin{figure}[H]
    \centering
    \includegraphics[width=0.65\columnwidth]{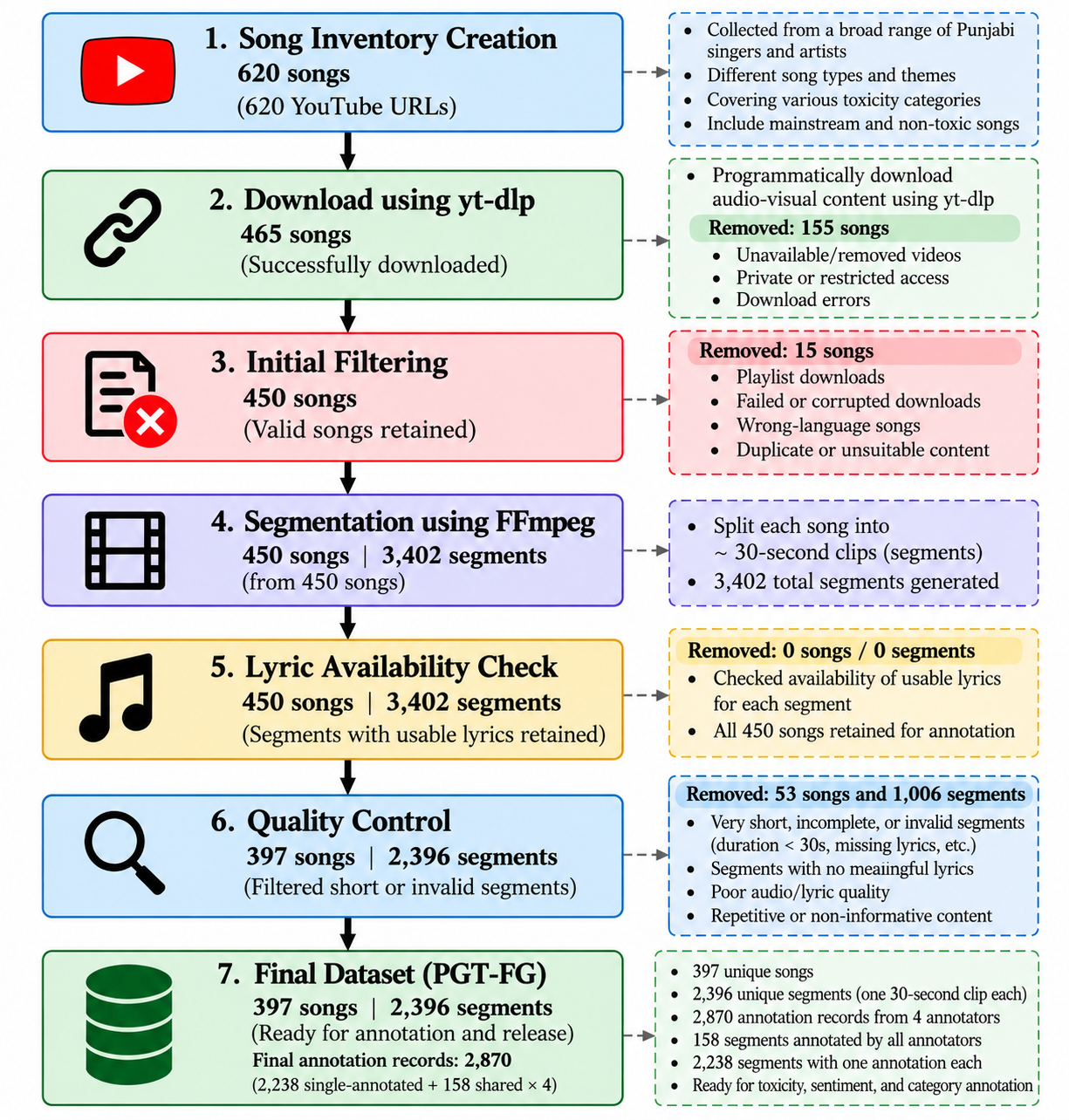}
  \caption{Dataset construction and quality-control pipeline. The initial
inventory contained 620 songs, of which 465 were successfully downloaded and
450 were retained after initial filtering. These songs produced 3,402
candidate segments. A lyric-availability check retained all 450 songs and
3,402 segments. Subsequent data-cleaning and multimodal quality-control
procedures removed 53 songs and 1,006 segments, resulting in a final dataset
of 397 unique songs and 2,396 unique segments. The final annotation collection
contains 2,870 original annotation records because 158 shared segments were
independently annotated by four annotators.}

    \label{fig:data_collection_pipeline}
\end{figure}

Figure~\ref{fig:data_collection_pipeline} illustrates the overall data
collection and preprocessing pipeline used to construct the dataset.

\subsection{Song Selection}

We selected songs to obtain broad coverage of different types of content
represented in Punjabi music, including songs containing potentially relevant
themes such as violence, misogyny, casteism, religious content, drug
references, gang references, profanity, and other potentially toxic content.
We also included mainstream and relatively non-toxic songs to provide
examples of both toxic and non-toxic content.

For each selected song, we collected and recorded its publicly available
YouTube URL as part of the dataset metadata. We subsequently used these URLs
to download the corresponding media for preprocessing and segmentation. The
selection process therefore focused on obtaining diversity in artists,
songs, themes, and content rather than relying exclusively on songs that
had already been identified as toxic.

The collection contains songs from a broad range of Punjabi singers and
artists. The final collection consists of 397 unique songs. This diversity
reduces dependence on the style, vocabulary, or lyrical themes of a small
number of artists and provides broader coverage of Punjabi musical content.

\subsection{YouTube Data Acquisition}

We collected the selected YouTube URLs and used \texttt{yt-dlp} to download
the corresponding audio-visual content. The download process was performed
programmatically to maintain consistency across the collected songs. During
collection, some URLs resulted in playlists or other collections of multiple
videos rather than individual songs. We identified these playlist-level
downloads during subsequent inspection and removed them because the unit of
analysis in the proposed resource is an individual song.

We excluded videos when the download failed, the source was no longer
available, or the downloaded content did not correspond to the intended
Punjabi song. We also removed songs with an incorrect language or unsuitable
content during quality checking. These filtering and inspection steps resulted in 465 successfully downloaded songs, of which 450 individual songs were retained for
segmentation. Playlist-level downloads, invalid or unsuitable sources,
and other non-individual-song content were excluded before segmentation.

\subsection{Multimodal Content and Quality Control}

The collected songs were intended to provide complementary text, audio, and
visual information for each segment. However, because the data were obtained
from naturally occurring YouTube content, the availability and quality of each
modality varied across songs. Most of the selected videos contained music-video
visuals, while a small number contained lyric-based visuals, static images,
thumbnails, or limited visual information. These variations were retained when
the content was otherwise suitable for the study.

The initial segmentation produced 3,402 candidate segments from 450 songs.
Before annotation, a lyric-availability check was performed to determine
whether usable lyrical content was available for the segmented material. No
songs or segments were removed during this preliminary check; therefore, all
450 songs and 3,402 candidate segments proceeded to the subsequent
quality-control stage.

The 3,402 candidate segments were then subjected to multimodal quality-control
and data-cleaning procedures before annotation. These procedures considered
the availability and quality of lyrics, audio, and visual content. Segments
corresponding to very short or incomplete clips, segments without meaningful
lyric content, poor audio or lyric quality, repetitive or non-informative
content, and other invalid material were excluded. A total of 53 songs and
1,006 segments were removed during this quality-control stage. After these
filtering procedures, 2,396 valid segments from 397 unique songs were retained
for annotation.

\subsection{Temporal Segmentation}

After downloading and initial quality checking, the 450 retained songs were
divided into approximately 30-second temporal segments using FFmpeg. A
consistent segment length was adopted to provide manageable and comparable
units for annotation and multimodal feature extraction. Each segment was
associated with its corresponding temporal interval in the original song.

For each segment, metadata including the song identifier, segment number,
start time, end time, artist name, and song title were maintained. The
corresponding lyrics were aligned with the same temporal segment wherever
lyrics were available. This temporal organization enables the text, audio, and
visual modalities to be analysed together for the same portion of a song.

The segmentation process generated 3,402 candidate segments from 450 songs.
A subsequent quality-control stage removed 1,006 unsuitable segments and
53 songs based on criteria including very short or incomplete segments,
missing or non-meaningful lyrics, poor audio or lyric quality, repetitive
content, and other invalid material. After these filtering steps, the final
dataset contained 2,396 valid segments from 397 unique songs.

\subsection{Ethical and Copyright Considerations}

The songs and music videos used in this study are publicly available
copyrighted works. The downloaded media were used temporarily for research processing,
including segmentation, annotation, and feature extraction. To avoid
redistributing copyrighted audio and video files, the released dataset
contains the corresponding YouTube URLs and research metadata rather than
the original media files.

The dataset is intended solely as a research resource for studying
multimodal toxicity in Punjabi musical content. The authors do not claim
ownership of the original songs, recordings, videos, or lyrics. Users of
the dataset are expected to access the original content through the
provided sources and comply with applicable copyright laws and platform
terms of service.

The purpose of the dataset is to analyse content-level characteristics such
as toxicity and sentiment and not to assess individual singers or artists.
Accordingly, artist information is treated as metadata associated with the
source material rather than as a basis for evaluating the artists
themselves.

\section{Annotation Methodology}
\label{sec:annotation}

The annotation methodology was designed to capture multiple dimensions of
potentially harmful content in Punjabi song lyrics while maintaining
consistency across annotators. Following the segmentation and quality-control
procedures described in Section~3.3, 2,396 valid segments from 397 unique
songs were retained for annotation. Each segment represented an
approximately 30-second portion of a song and was associated with its
available lyric transcript, video, audio context, and song metadata. The
annotation process considered the lyric transcript together with the
corresponding video and audio context. Each segment received labels for
overall toxicity, fine-grained toxicity categories, severity, sentiment,
mode, and annotation uncertainty.

The retained segments were distributed among four Punjabi-native speaker
annotators. Each segment was assigned to a designated annotator, while an
initial shared subset of 200 segments was additionally selected for
independent annotation by all four annotators to assess inter-annotator
agreement. After annotation matching and completeness checks, 158 shared
segments had complete annotations from all four annotators and were retained
for inter-annotator agreement analysis. The final annotation collection
contains 2,870 annotation records: 2,238 segments with one annotation record
each and 158 shared segments with four annotation records each.

The annotation process consisted of four main stages: (i) annotator
preparation and guideline familiarization, (ii) independent annotation of
the quality-controlled song segments, (iii) measurement of
inter-annotator agreement using the shared subset, and (iv) derivation of
a single resolved label per segment for dataset-level statistics. The
following subsections describe these stages and the annotation scheme in
detail.

\subsection{Annotation Team and Preparation}

Four Punjabi-native speaker annotators participated in the annotation
process. The annotators were selected based on their familiarity with the
Punjabi language and their ability to understand Punjabi--English
code-mixed lyrics. No personal identifying information about the annotators
is reported in this paper.

Before beginning the main annotation task, all annotators reviewed a shared
annotation guideline document that defined the label categories, coding
values, and decision rules. Particular attention was given to distinguishing
toxic and non-toxic content, assigning multiple toxicity categories when
necessary, and applying severity and mode consistently. The annotators also
reviewed example segments to establish a common understanding of the
annotation criteria.

A pilot annotation round was conducted before the main annotation stage.
The pilot allowed the annotators to familiarize themselves with the
annotation interface and identify cases in which the guidelines required
clarification. The annotation guidelines were subsequently used as the
common reference throughout the main annotation process.

\subsection{Annotation Unit and Workflow}

The basic annotation unit was a lyric segment corresponding to an
approximately 30-second portion of a song. The final annotation pool
consisted of 2,396 quality-controlled segments from 397 unique songs. Each
segment was associated with its corresponding video, audio interval,
available transcript, and song metadata. Annotators reviewed the available
multimodal information before assigning the labels.

For each assigned segment, annotators independently examined the available
transcript and video and considered the audio context when necessary.
Annotations were performed at the segment level rather than at the
complete-song level. This allowed different portions of the same song to
receive different labels when the lyrical content changed over time.

Each annotator performed the annotation independently without access to the
decisions of the other annotators. This independent annotation procedure
was used to reduce the possibility of one annotator influencing another
annotator's judgment. For every assigned segment, the annotators recorded
values for overall toxicity, fine-grained toxicity categories, severity,
sentiment, mode, and uncertainty. The shared subset was independently
annotated by the participating annotators and was subsequently used for
inter-annotator agreement analysis.

\subsection{Label Schema}

The annotation scheme consists of six complementary label types: toxicity,
fine-grained toxicity categories, severity, sentiment, mode, and uncertainty
flag. Table~\ref{tab:label_schema} summarizes the encoding used for each
label.

\begin{table}[htbp]
\centering
\caption{Label schema used for annotation of PGT-FG.}
\label{tab:label_schema}

\small
\resizebox{\columnwidth}{!}{%
\begin{tabular}{|l|c|l|}
\hline
\rowcolor{blue!15}
\textbf{Label} & \textbf{Type} & \textbf{Allowed Values and Meaning} \\
\hline

Toxicity & Binary &
0 = Non-toxic; 1 = Toxic \\
\hline

Categories & Multi-label &
Violence, Misogyny, Casteism, Religious, Drug,
Gang References, Profanity, Other \\
\hline

Severity & Ordinal &
$-1$ = N/A; 0 = Mild; 1 = Moderate; 2 = Severe \\
\hline

Sentiment & Categorical &
0 = Negative; 1 = Neutral; 2 = Positive \\
\hline

Mode & Categorical &
$-1$ = N/A; 0 = Explicit; 1 = Implicit; 2 = Mixed \\
\hline

Uncertainty Flag & Binary &
0 = No uncertainty flagged; 1 = Uncertainty flagged \\
\hline

\end{tabular}%
}
\end{table}

\subsection{Toxicity and Fine-Grained Categories}

The primary toxicity label is binary, with 0 representing non-toxic content
and 1 representing toxic content. A segment was labelled as toxic when its
content contained harmful, abusive, degrading, threatening, or otherwise
potentially harmful material according to the annotation guidelines.

For toxic segments, annotators additionally assigned one or more fine-grained
toxicity categories. The categories were treated as multi-label annotations
because a single lyric segment may contain more than one form of harmful
content. The eight categories were violence, misogyny, casteism, religious
content, drug references, gang references, profanity, and other toxic
content.

\paragraph{Violence.}
This category covers references to physical violence, violent acts, or
threats of physical harm. Examples include lyrics describing killing,
shooting, beating, attacking, or explicitly threatening physical harm.

\paragraph{Misogyny.}
This category covers derogatory, degrading, objectifying, or hostile
representations of women. Examples include lyrics that demean women because
of their gender, reduce women to sexual objects, or express gender-based
contempt.

\paragraph{Casteism.}
This category covers derogatory or discriminatory references to caste groups,
including caste-based insults, humiliation, exclusion, or claims of
superiority or inferiority based on caste.

\paragraph{Religious Content.}
This category covers harmful or derogatory references directed toward
religions, religious groups, religious figures, or religious identities.
Examples include religiously targeted insults, derogatory statements, or
content promoting hostility toward a religious group.

\paragraph{Drug References.}
This category covers references to illegal or harmful drug use, possession,
distribution, or glorification. Examples include lyrics explicitly
celebrating drug consumption or describing drug-related activities.

\paragraph{Gang References.}
This category covers references to gangs, criminal groups, gang identity,
criminal-group affiliation, or the glorification of gang-related activity.

\paragraph{Profanity.}
This category covers explicit vulgar, obscene, or profane language. A
segment may receive this label even when the profanity does not independently
constitute another toxicity category.

\paragraph{Other Toxic Content.}
This category captures toxic or harmful content that does not fit the
specific categories above. It provides coverage for cases that may involve
other forms of harmful expression without forcing them into an inappropriate
category.

\subsection{Severity}

Severity was assigned to characterize the intensity of toxic content. The
annotation scheme uses four codes: $-1$ for non-toxic or not applicable,
0 for mild, 1 for moderate, and 2 for severe toxicity.

Mild toxicity refers to relatively limited harmful content, such as a single
profanity or a brief mention of harmful behaviour. Moderate toxicity includes
more sustained forms of harmful expression, such as glorification,
objectification, or general threats. Severe toxicity represents explicit and
high-intensity harmful content, including explicit violence, named targets,
graphic content, or incitement.

The $-1$ value was used only when severity was not applicable, particularly
for non-toxic segments. This distinction prevents a missing value from being
confused with a genuine severity level.

\begin{table}[htbp]
\centering
\caption{Decision criteria for severity and mode annotation.}
\label{tab:severity_mode}

\resizebox{\columnwidth}{!}{%
\begin{tabular}{|l|l|}
\hline
\rowcolor{blue!15}
\textbf{Label} & \textbf{Decision Criterion} \\
\hline

Severity = $-1$ &
Non-toxic or not applicable \\
\hline

Severity = 0 &
Mild content, such as a single profanity or a brief mention of harmful content \\
\hline

Severity = 1 &
Moderate content, including sustained glorification,
objectification, or general threats \\
\hline

Severity = 2 &
Severe content, including explicit violence, named targets,
graphic content, or incitement \\
\hline

Mode = $-1$ &
Not applicable \\
\hline

Mode = 0 &
Explicit harmful content stated directly \\
\hline

Mode = 1 &
Implicit harmful meaning requiring interpretation or contextual understanding \\
\hline

Mode = 2 &
A combination of explicit and implicit harmful content \\
\hline

\end{tabular}%
}
\end{table}

\subsection{Sentiment}

Each segment was also assigned a sentiment label independently of the
toxicity label. Sentiment was encoded categorically as 0 for Negative,
1 for Neutral, and 2 for Positive. The sentiment label captures the overall
emotional orientation of the segment and is therefore distinct from
toxicity. A segment can, for example, express a positive sentiment while
still containing toxic content, or express negative sentiment without being
toxic.

\subsection{Mode and Uncertainty}

The mode label captures how toxic content is expressed in the lyrical text. A
value of 0 indicates that the toxicity is stated explicitly, such as through
direct harmful language, named targets, or overt threats. A value of 1 indicates
implicit toxicity, which requires interpretation of context, metaphor, or
culturally specific references. A value of 2 indicates mixed cases in which both
explicit and implicit forms of toxicity are present in the same segment. For
non-toxic segments, mode is set to $-1$ to indicate that the label is not
applicable. This distinction supports the analysis of direct versus
context-dependent expressions of toxicity, including coded language, cultural
references, and euphemistic expressions.

The uncertainty flag captures whether an annotator considered a segment
uncertain during annotation. A value of 0 indicates that no uncertainty was
flagged, while a value of 1 indicates that the annotator flagged the segment as
uncertain. The flag was included to identify potentially ambiguous segments and
support subsequent analysis of difficult annotation cases.

\subsection{Annotation Rules and Consistency}

Several consistency rules were applied throughout the annotation process.
First, toxicity categories were treated as multi-label fields, allowing
multiple categories to be active for the same segment. Second, every toxic
segment was required to have at least one active toxicity category. Third,
when a segment was labelled non-toxic, all toxicity category fields were set
to zero, while severity and mode were assigned $-1$ to indicate that they
were not applicable.

The use of explicit numerical codes for non-applicable fields was adopted
instead of leaving these fields blank. This distinction is important for
subsequent computational processing because it separates an intentional
non-applicable value from missing annotation data. The uncertainty flag was encoded consistently as a binary field, with 0
indicating that no uncertainty was flagged and 1 indicating that uncertainty
was flagged. The same coding scheme
was used across the annotation files to ensure consistency during dataset
integration and analysis.

\subsection{Inter-Annotator Agreement}
\label{sec:inter_annotator_agreement}

To assess the consistency of the annotation process, a shared subset of the
dataset was independently annotated by all four annotators. After matching
the annotation records using the unique segment identifier
(\texttt{verse\_id}), 158 segments with complete annotations from all four
annotators were identified, resulting in 632 annotation records. Fleiss'
Kappa was calculated separately for each annotation field.

Table~\ref{tab:fleiss_kappa} reports the agreement obtained for the major
annotation fields. The primary toxicity label achieved a Kappa of
$\kappa=0.6521$, indicating substantial agreement. Among the fine-grained
toxicity categories, profanity showed the highest agreement
($\kappa=0.8430$), followed by drug references ($\kappa=0.7488$), misogyny
($\kappa=0.6651$), and violence ($\kappa=0.6354$). Moderate agreement was
observed for the other category ($\kappa=0.5725$) and gang references
($\kappa=0.4952$). For the additional annotation dimensions, severity showed moderate
agreement ($\kappa = 0.5680$), sentiment showed fair agreement
($\kappa = 0.2508$), and mode showed substantial agreement
($\kappa = 0.6145$).

\begin{table}[htbp]
\centering
\caption{Inter-annotator agreement measured using Fleiss' Kappa on the
shared subset of 158 segments independently annotated by four annotators.}
\label{tab:fleiss_kappa}

\scriptsize
\begin{tabular}{|l|c|l|}
\hline
\rowcolor{blue!15}
\textbf{Label} & \textbf{$\kappa$} & \textbf{Interpretation} \\
\hline
Toxicity & 0.6521 & Substantial \\
\hline
Violence & 0.6354 & Substantial \\
\hline
Misogyny & 0.6651 & Substantial \\
\hline
Casteism & -0.0016 & Poor \\
\hline
Religious & N/A & Not estimable -- no variation \\
\hline
Drug & 0.7488 & Substantial \\
\hline
Gang References & 0.4952 & Moderate \\
\hline
Profanity & 0.8430 & Almost perfect \\
\hline
Other & 0.5725 & Moderate \\
\hline
Severity & 0.5680 & Moderate \\
\hline
Sentiment & 0.2508 & Fair \\
\hline
Mode & 0.6145 & Substantial \\
\hline
\end{tabular}
\end{table}

The very low agreement observed for casteism should be interpreted in the
context of the limited occurrence and variation of this category in the
shared subset. The casteism category was represented by very few positive
examples, which makes category-specific agreement estimates less stable and
more sensitive to individual annotation decisions. In addition, caste-related
content may require contextual interpretation to distinguish discriminatory
expressions from references to cultural or social identity. The casteism
category therefore yielded a Kappa of $\kappa=-0.0016$, indicating agreement
approximately at the chance level for the observed ratings. Similarly, the
religious category was not estimable because there was no variation in the
ratings within the shared subset. This result should not be interpreted as
evidence of disagreement; rather, Fleiss' Kappa cannot be estimated when the
ratings contain no variation.

These outcomes are consistent with the low frequency of casteism and
religious labels in the full dataset, as reported in
Section~\ref{sec:dataset_statistics}. They highlight a limitation of the
current dataset for evaluating the rarest fine-grained toxicity categories
and motivate future expansion of these categories with additional annotated
examples. Nevertheless, the substantial agreement obtained for the primary
toxicity label and several of the more frequently represented fine-grained
categories provides evidence that the annotation procedure can produce
reasonably consistent labels for the major toxicity dimensions.

\subsection{Derivation of Segment-Level Labels for Dataset Statistics}

The complete annotation records were retained to preserve the individual
decisions of the annotators. However, dataset-level statistics require
one label per unique segment. Therefore, a separate resolved
representation was constructed without modifying the original annotation
records.

The final annotation data contain 2,396 unique segments. Of these, 2,238
segments have a single annotation, whereas 158 segments have complete
annotations from all four annotators. 
Thus, the complete annotation collection contains

\[
2,238 + (158 \times 4) = 2,870
\]

annotation records.

For singly annotated segments, the original annotation was retained. For
the 158 shared segments, the toxicity label was resolved using majority
voting. In the event of a 2--2 tie, the toxic label was selected as the
more conservative outcome. Fine-grained toxicity categories were resolved
using the annotations of annotators who marked the segment as toxic, with
majority voting applied independently to each category. Binary ties were
resolved in favour of the positive category.

Severity was resolved using majority voting among the toxic annotators,
with the higher severity selected in the event of a tie. Mode was resolved
using majority voting among the toxic annotators, with a deterministic
annotator-order rule used for unresolved ties. Sentiment was resolved using
majority voting among the four annotators, with the predetermined
annotator-order rule used to resolve ties. For non-toxic segments,
toxicity categories were set to zero and severity and mode were set to
$-1$.
For the uncertainty flag, a shared segment was assigned a value of 1 if at
least one of the four annotators flagged uncertainty; otherwise, the resolved
value was 0.

The resulting one-row-per-segment representation was used for the
segment-level descriptive statistics reported in Section~5, while the
complete annotation records were retained for transparency and
inter-annotator agreement analysis.

\subsection{Ethical Considerations and Annotator Wellbeing}

The annotation task involved exposure to potentially offensive, violent,
sexually degrading, discriminatory, and otherwise harmful lyrical content.
The potentially sensitive nature of the material was considered when
organizing the annotation process.

Annotators were informed about the nature of the material involved in the
task. No personal identifying information about the annotators is included
in the dataset or reported in this paper. The annotation process was
conducted for research purposes, and the released resource contains
annotations and metadata rather than redistributed copyrighted audio or
video content.

\subsection{Summary of the Annotation Process}

Overall, the annotation methodology combines independent annotation,
fine-grained toxicity categories, severity assessment, sentiment analysis,
mode identification, and explicit uncertainty reporting. The use of
standardized coding rules provides a consistent framework for assigning
labels across the dataset. A shared subset of 158 segments was independently
annotated by all four annotators and was used to assess inter-annotator
agreement using Fleiss' Kappa.

The individual annotation records were retained to support
reproducibility and inter-annotator agreement analysis. For the final
dataset-level statistics, the 2,238 singly annotated segments retained
their original labels, while the 158 shared segments were resolved using
the rules described in Section~4.10. This resolution produced one
segment-level label representation containing 2,396 unique segments.

\section{Dataset Statistics and Distribution}
\label{sec:dataset_statistics}

This section provides an overview of the composition and distribution of PGT-FG. The final dataset contains 397 unique songs and
2,396 unique song segments. Each segment represents an approximately
30-second portion of a song and is associated with its corresponding lyric
transcript, audio, video, and annotation metadata. The dataset provides
binary toxicity labels together with fine-grained toxicity categories,
severity, sentiment, mode, and annotation uncertainty.

\subsection{Dataset Composition}

The final dataset consists of 397 unique songs divided into 2,396 unique segments. The segments were obtained by dividing the collected songs into approximately 30-second windows. Very short or incomplete segments were removed during preprocessing. The resulting dataset therefore provides segment-level annotations while retaining the temporal context of the original songs. The unique-segment count of 2,396 is therefore used as the basis for all
subsequent dataset-level statistics, with the resolved one-row-per-segment
representation used to ensure that shared segments are counted only once.

\begin{table}[htbp]
\centering
\caption{Overall statistics of PGT-FG.}
\label{tab:dataset_statistics}

\begin{tabular}{|l|r|}
\hline
\rowcolor{blue!15}
\textbf{Statistic} & \textbf{Value} \\
\hline
Total Songs Collected & 620 \\
\hline
Successfully Downloaded Songs & 465 \\
\hline
Initial Segments & 3,402 \\
\hline
Final Unique Songs & 397 \\
\hline
Final Unique Segments & 2,396 \\
\hline
Annotation Records & 2,870 \\
\hline
Total Audio Duration & $\approx$20 hours \\
\hline
\end{tabular}
\end{table}

The dataset contains 2,396 unique segments but 2,870 total annotation
records. The difference arises because 158 segments in the shared subset
were independently annotated by all four annotators, resulting in
$158 \times 4 = 632$ annotation records for the shared subset. The remaining
2,238 segments have one annotation record each. Therefore, the total number
of annotation records is $2,238 + 632 = 2,870$. The unique-segment count
of 2,396 is used as the basis for all subsequent dataset-level statistics
to avoid counting the shared segments multiple times.

\subsection{Toxicity Distribution}

The final dataset contains 2,396 unique song segments, of which 1,691
(70.58\%) are labelled as non-toxic and 705 (29.42\%) are labelled as
toxic. Thus, approximately 29\% of the segments contain potentially toxic
content, while the remaining segments are classified as non-toxic. The
distribution of the binary toxicity labels is presented in
Table~\ref{tab:toxicity_distribution}.

\begin{table}[H]
\centering
\caption{Distribution of toxic and non-toxic segments.}
\label{tab:toxicity_distribution}
\small
\begin{tabular}{|l|r|r|}
\hline
\rowcolor{blue!15}
\textbf{Toxicity} & \textbf{Count} & \textbf{Percentage} \\
\hline
Non-toxic & 1,691 & 70.58\% \\
\hline
Toxic & 705 & 29.42\% \\
\hline
\textbf{Total} & \textbf{2,396} & \textbf{100.00\%} \\
\hline
\end{tabular}
\end{table}

\begin{figure}[H]
\centering
\includegraphics[width=0.78\columnwidth]{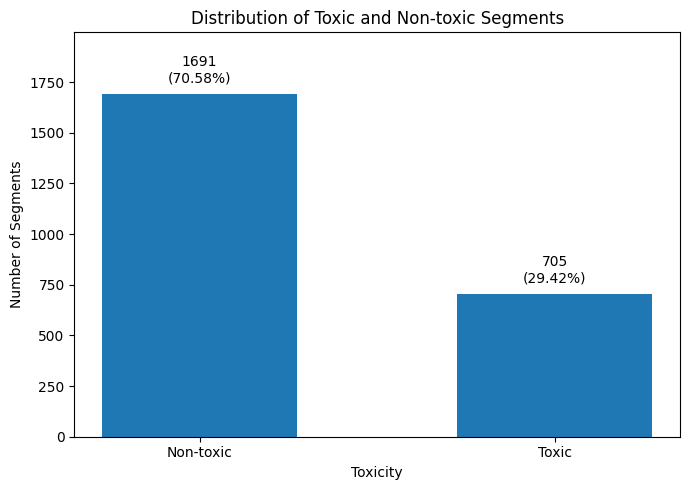}
\caption{Distribution of toxic and non-toxic segments in the final dataset.}
\label{fig:toxicity_class_balance}
\end{figure}

\subsection{Severity, Sentiment, and Mode Distribution}

Severity and mode were applicable only to toxic segments. Among the 705 toxic
segments, 112 (15.89\%) were classified as mild, 357 (50.64\%) as moderate,
and 236 (33.48\%) as severe. The remaining 1,691 segments are non-toxic and
therefore have severity marked as non-applicable ($-1$). Thus, the dataset
contains $705 + 1,691 = 2,396$ unique segments in total.

The mode annotation was also applicable only to toxic segments. Among the
705 toxic segments, 567 (80.43\%) were classified as explicit, 106 (15.04\%)
as implicit, and 32 (4.54\%) as mixed. Thus, the mode distribution also
accounts for all 705 toxic segments.

The sentiment annotation was applied at the segment level. Across the
2,396 unique segments, 425 (17.74\%) were labelled negative, 395 (16.49\%)
were labelled neutral, and 1,576 (65.78\%) were labelled positive.

\begin{figure}[H]
\centering
\includegraphics[width=0.78\columnwidth]{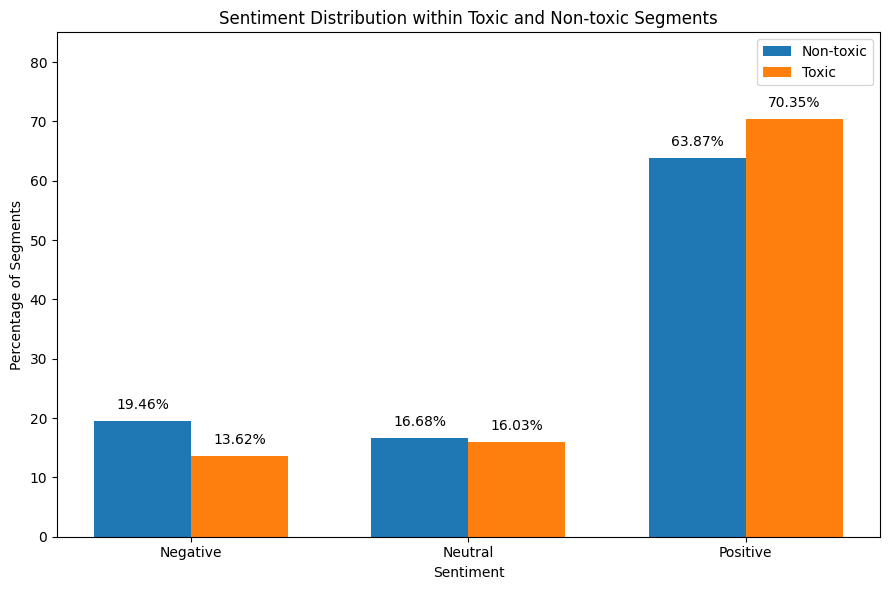}
\caption{Sentiment distribution within the toxic and non-toxic segments.
Positive sentiment was the dominant category in both groups, accounting for
63.87\% of non-toxic segments and 70.35\% of toxic segments. Negative
sentiment accounted for 19.46\% of non-toxic and 13.62\% of toxic segments,
while neutral sentiment accounted for 16.68\% and 16.03\%, respectively.
The distribution illustrates that sentiment and toxicity are distinct
annotation dimensions and that toxic content can occur within segments
expressing an overall positive sentiment.}
\end{figure}

\begin{table}[H]
\centering
\caption{Severity distribution of toxic segments.}
\label{tab:severity_distribution}
\small
\begin{tabular}{|l|r|r|}
\hline
\rowcolor{blue!15}
\textbf{Severity} & \textbf{Count} & \textbf{Percentage} \\
\hline
Mild & 112 & 15.89\% \\
\hline
Moderate & 357 & 50.64\% \\
\hline
Severe & 236 & 33.48\% \\
\hline
\textbf{Total} & \textbf{705} & \textbf{100.00\%} \\
\hline
\end{tabular}
\end{table}

\begin{table}[H]
\centering
\caption{Sentiment distribution of unique song segments.}
\label{tab:sentiment_distribution}
\small
\begin{tabular}{|l|r|r|}
\hline
\rowcolor{blue!15}
\textbf{Sentiment} & \textbf{Count} & \textbf{Percentage} \\
\hline
Negative & 425 & 17.74\% \\
\hline
Neutral & 395 & 16.49\% \\
\hline
Positive & 1,576 & 65.78\% \\
\hline
\textbf{Total} & \textbf{2,396} & \textbf{100.00\%} \\
\hline
\end{tabular}
\end{table}

\begin{table}[H]
\centering
\caption{Mode distribution of toxic segments.}
\label{tab:mode_distribution}
\small
\begin{tabular}{|l|r|r|}
\hline
\rowcolor{blue!15}
\textbf{Mode} & \textbf{Count} & \textbf{Percentage} \\
\hline
Explicit & 567 & 80.43\% \\
\hline
Implicit & 106 & 15.04\% \\
\hline
Mixed & 32 & 4.54\% \\
\hline
\textbf{Total} & \textbf{705} & \textbf{100.00\%} \\
\hline
\end{tabular}
\end{table}

\subsection{Fine-Grained Toxicity Distribution}
\label{sec:fine_grained_toxicity}

In addition to the binary toxicity label, each toxic segment was assigned
one or more fine-grained toxicity categories. The annotation scheme
included violence, misogyny, casteism, religious content, drug
references, gang references, profanity, and other forms of toxicity.
The categories were treated as multi-label annotations, allowing a
single toxic segment to belong to more than one category.

The final dataset contains 705 unique toxic segments. Among these,
violence was the most frequently observed category, occurring in
539 segments (76.45\%). Gang references were identified in
227 segments (32.20\%), followed by drug-related content in
133 segments (18.87\%) and profanity in 62 segments (8.79\%).
The ``other'' category was observed in 19 segments (2.70\%),
while misogyny was identified in 20 segments (2.84\%). Casteism
was identified in 2 segments (0.28\%), and no segment was assigned
to the religious category.

\begin{figure}[H]
    \centering
    \includegraphics[width=\columnwidth]{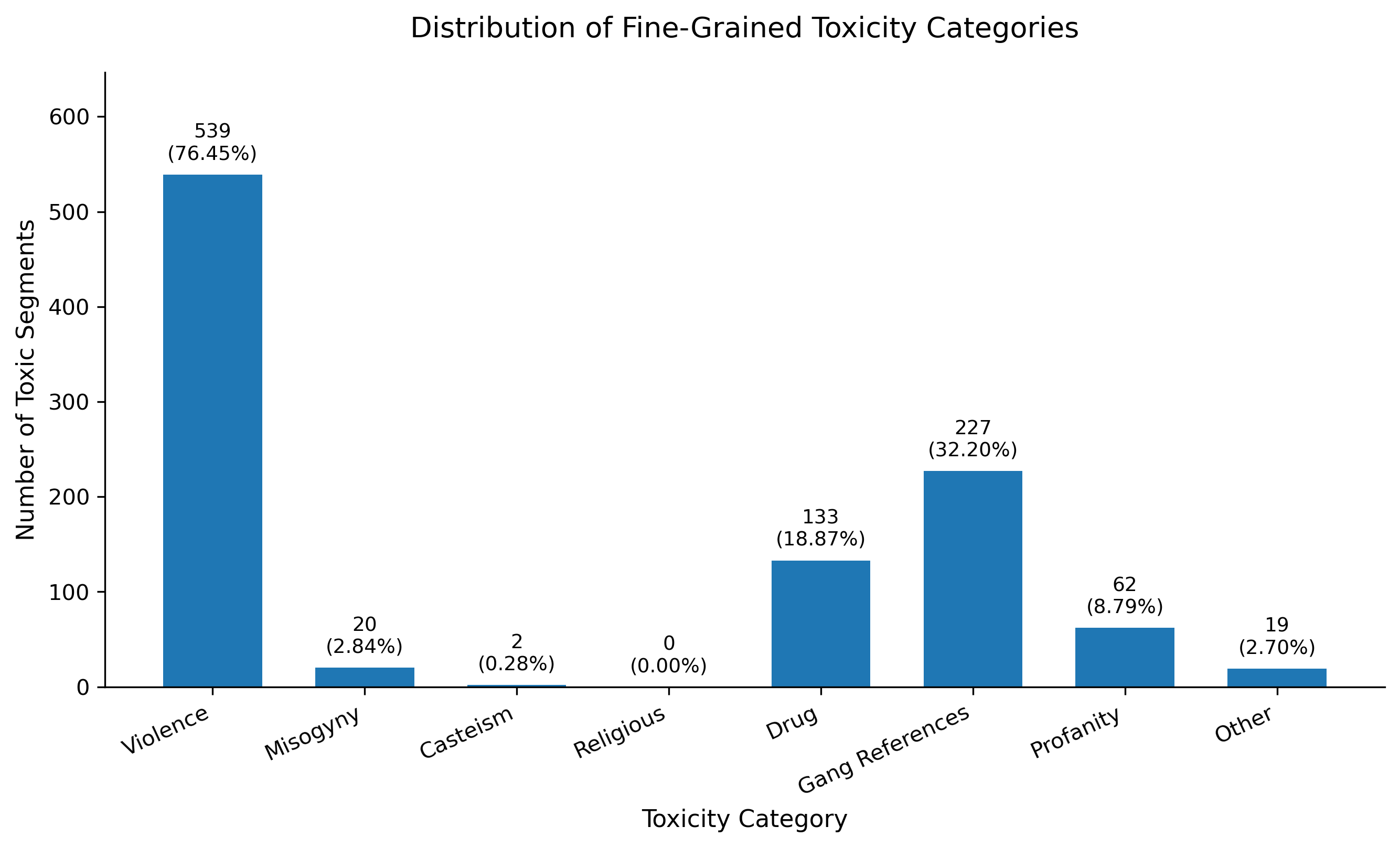}
    \caption{Distribution of fine-grained toxicity categories among the
705 toxic segments in the final dataset. Since the categories are
multi-label, a segment may belong to more than one category; therefore,
the percentages do not sum to 100\%.}
    \label{fig:toxicity_categories}
\end{figure}

The complete distribution of the fine-grained toxicity categories is
presented in Table~\ref{tab:toxicity_categories} and
Figure~\ref{fig:toxicity_categories}. Since the annotation scheme is
multi-label, a segment may belong to more than one toxicity category.
Consequently, the category percentages do not sum to 100\%.

\begin{table}[htbp]
\centering
\caption{Distribution of fine-grained toxicity categories among toxic segments.}
\label{tab:toxicity_categories}

\small
\begin{tabular}{|l|r|r|}
\hline
\rowcolor{blue!15}
\textbf{Toxicity Category} & \textbf{Count} & \textbf{Percentage} \\
\hline
Violence & 539 & 76.45\% \\
\hline
Misogyny & 20 & 2.84\% \\
\hline
Casteism & 2 & 0.28\% \\
\hline
Religious & 0 & 0.00\% \\
\hline
Drug & 133 & 18.87\% \\
\hline
Gang References & 227 & 32.20\% \\
\hline
Profanity & 62 & 8.79\% \\
\hline
Other & 19 & 2.70\% \\
\hline
\end{tabular}

\vspace{2pt}

\parbox{\columnwidth}{\footnotesize
\textit{Note:} The toxicity categories are multi-label; therefore,
a segment may belong to more than one category and the percentages
do not sum to 100\%.
}

\end{table}

Figure~\ref{fig:toxicity_categories} illustrates the distribution of
the fine-grained toxicity categories across the toxic segments. The
dominance of violence-related content is evident, while gang
references, drug-related content, and profanity constitute additional
substantial categories. The remaining categories occur less frequently
in the final dataset.

\subsection{Summary}

Overall, the dataset contains 397 unique songs and 2,396 unique segments,
with each segment representing approximately 30 seconds of multimodal song
content. Of these segments, 705 are labelled as toxic and 1,691 as
non-toxic. The distribution of toxicity, fine-grained toxicity categories,
severity, sentiment, and mode provides multiple levels of annotation for
studying harmful content in Punjabi music. The multi-label category scheme
also allows a segment to represent multiple forms of toxicity simultaneously,
making the dataset suitable for both binary toxicity detection and
fine-grained analysis of harmful content. The combination of segment-level
toxicity, sentiment, severity, and mode annotations supports both
single-task and multi-task modelling approaches and can facilitate future
research on multimodal toxicity analysis.

\section{Data Access and Format}
\label{sec:data_access}

PGT-FG is publicly available through Zenodo, an open-access research data
repository. The release provides the research-generated annotation data,
associated segment and song metadata, and documentation describing the
dataset structure and usage. Original lyric text, lyric excerpts, and
copyrighted audio or video content are not redistributed in the public
release. No source code is included in the current release. The dataset is
available at
\url{https://doi.org/10.5281/zenodo.23040156}.

The primary released annotation file is
\texttt{PGT-FG\_annotations\_v1.csv}, provided in comma-separated values
(CSV) format. The file contains one final resolved annotation for each
unique song segment. It contains 21 fields describing the segment
identifier, source-song metadata, temporal boundaries, source YouTube URL,
binary toxicity label, fine-grained toxicity categories, severity, sentiment,
mode, and the binary uncertainty flag. The complete schema of the released
annotation file is provided in Table~\ref{tab:released_csv_schema}. The
\texttt{verse\_id} uniquely identifies each song segment, whereas
\texttt{song\_id} identifies the corresponding source song.

The final dataset contains 2,396 unique segments from 397 unique songs.
The original annotation process produced 2,870 individual annotation records
from four annotators. Of the 2,396 unique segments, 158 were independently
annotated by all four annotators, resulting in $158 \times 4 = 632$
annotation records. The remaining 2,238 segments received one annotation
record each. Thus, the total number of original annotation records was
$2,238 + 632 = 2,870$. The labels for the 158 shared segments were resolved
using the majority-voting procedure described in Section~4.10.
Consequently, the released annotation file contains one final resolved
record per segment rather than the original repeated annotation records.

The released resource consists of three main components. First, the primary
annotation file, \texttt{PGT-FG\_annotations\_v1.csv}, contains the final
segment-level annotations described in Table~\ref{tab:released_csv_schema}.
These annotations constitute the primary research contribution of the
dataset. Each row corresponds to one unique song segment and contains the
resolved toxicity labels together with the associated metadata.

Second, segment and source-song metadata are provided through
\texttt{segment\_metadata.csv} and \texttt{song\_metadata.csv}. The segment
metadata include the segment identifier, source-song identifier, segment
start and end times, and segment duration. The song metadata include the
song identifier, source video identifier, song or video title, associated
artist, source YouTube URL, and source media duration. Researchers may use
the identifiers, timestamps, and source references to locate and process the
corresponding source material independently, subject to its continued
availability and the applicable copyright and platform terms.

Third, the Zenodo release includes documentation in the form of
\texttt{README.md} and \texttt{DATA\_DICTIONARY.md}. The README provides an
overview of the dataset, dataset statistics, source-media and copyright
considerations, and general usage information. The data dictionary describes
the structure and fields of the released annotation and metadata files.
No source code or annotator-level raw annotation file is included in the
current Zenodo release.

The released annotation file does not contain the original lyric text or
other lyric-derived text fields. In particular, the
\texttt{text\_segment}, \texttt{text\_original}, and
\texttt{text\_clean} fields used during the annotation and preprocessing
stages are not redistributed in the primary public annotation file. No
lyric excerpts are included in the Zenodo release. This design avoids
redistributing copyrighted lyric content while retaining the segment-level
annotations and source references required for research use. Researchers
who require the corresponding lyrical material must obtain it independently
from the relevant lawful sources and are responsible for complying with
applicable copyright and platform terms.

The dataset does not include or redistribute the original copyrighted audio
or video recordings of the source songs. No raw audio or video files are
included in the Zenodo release package. Researchers who require the original
audio-visual material must obtain it independently from the identified
public sources. The dataset authors do not claim ownership of the original
songs, sound recordings, videos, lyrics, or other copyrighted source
materials. The availability of the original source material may change over
time if videos are removed, restricted, or made private.

The annotation data and research-generated metadata provided by the authors
are released under the Creative Commons Attribution 4.0 International
(CC BY 4.0) license. This license applies to the research annotations and
metadata provided by the authors and does not grant rights to any underlying
copyrighted audio, video, or lyric content obtained from third-party
sources. Users reusing the released annotations should provide appropriate
attribution to the dataset and its authors.

The released annotation file can be loaded using standard data-analysis
libraries. For example, the following Python code loads the released
annotation file and reports the number of records, unique segments, and
unique songs:

\begin{verbatim}
import pandas as pd

df = pd.read_csv("PGT-FG_annotations_v1.csv")

print("Annotation records:", len(df))
print("Unique segments:", df["verse_id"].nunique())
print("Unique songs:", df["song_id"].nunique())
\end{verbatim}

For the released file, the expected output is:

\begin{verbatim}
Annotation records: 2396
Unique segments: 2396
Unique songs: 397
\end{verbatim}

The distinction between the original annotation records and the final
resolved dataset is important when reproducing the dataset statistics
reported in this paper. The original annotation process generated 2,870
individual annotation records, whereas the released annotation file
contains 2,396 final segment-level records, with exactly one resolved
annotation per unique \texttt{verse\_id}. The original 2,870 annotator-level
records are retained as part of the internal annotation process but are not
included in the public Zenodo release.

\begin{table*}[htbp]
\centering
\caption{Schema of the released PGT-FG annotation CSV file.}
\label{tab:released_csv_schema}
\scriptsize
\begin{tabular}{|l|l|p{2.7cm}|p{5.2cm}|}
\hline
\rowcolor{blue!15}
\textbf{Field} & \textbf{Data Type} & \textbf{Allowed Values} &
\textbf{Description} \\
\hline

\texttt{verse\_id} & String &
Unique identifier &
Unique identifier assigned to each song segment. \\
\hline

\texttt{song\_id} & String &
Song identifier &
Identifier linking the segment to its source song. \\
\hline

\texttt{start\_sec} & Float &
$\geq 0$ &
Start time of the segment in seconds. \\
\hline

\texttt{end\_sec} & Float &
$\geq 0$ &
End time of the segment in seconds. \\
\hline

\texttt{duration\_sec} & Float &
$>0$ &
Duration of the segment in seconds. \\
\hline

\texttt{artist\_name} & String &
Artist name &
Name of the artist associated with the source song. \\
\hline

\texttt{title} & String &
Song title &
Title of the source song. \\
\hline

\texttt{toxicity} & Integer &
0, 1 &
0 = Non-toxic; 1 = Toxic. \\
\hline

\texttt{cat\_violence} & Integer &
0, 1 &
Indicates the presence of violence-related toxic content. \\
\hline

\texttt{cat\_misogyny} & Integer &
0, 1 &
Indicates the presence of misogynistic or gender-based toxic content. \\
\hline

\texttt{cat\_casteism} & Integer &
0, 1 &
Indicates the presence of caste-based discriminatory content. \\
\hline

\texttt{cat\_religious} & Integer &
0, 1 &
Indicates harmful or derogatory content targeting religious groups,
identities, or figures. \\
\hline

\texttt{cat\_drug} & Integer &
0, 1 &
Indicates drug-related toxic content or references. \\
\hline

\texttt{cat\_gang\_references} & Integer &
0, 1 &
Indicates gang, criminal-group, or gang-affiliation references. \\
\hline

\texttt{cat\_profanity} & Integer &
0, 1 &
Indicates explicit vulgar, obscene, or profane language. \\
\hline

\texttt{cat\_other} & Integer &
0, 1 &
Indicates toxic content not covered by the other defined categories. \\
\hline

\texttt{severity} & Integer &
$-1,0,1,2$ &
$-1$ = Not applicable; 0 = Mild; 1 = Moderate; 2 = Severe. \\
\hline

\texttt{sentiment} & Integer &
0, 1, 2 &
0 = Negative; 1 = Neutral; 2 = Positive. \\
\hline

\texttt{mode} & Integer &
$-1,0,1,2$ &
$-1$ = Not applicable; 0 = Explicit; 1 = Implicit; 2 = Mixed. \\
\hline

\texttt{uncertainty\_flag} & Integer &
0, 1 &
0 = No uncertainty flagged; 1 = Uncertainty flagged. \\
\hline

\texttt{youtube\_url} & String &
YouTube URL &
URL of the publicly available YouTube source video associated with the
source song. \\
\hline

\end{tabular}
\end{table*}

\section{Limitations and Future Work}
\label{sec:limitations_future_work}

Despite providing a multimodal and fine-grained resource for toxicity
analysis in Punjabi music, the dataset has several limitations.

First, the dataset contains 2,396 unique segments from 397 songs. Although
the dataset provides multiple annotation dimensions, its size limits the
coverage of the broader Punjabi music domain. The available songs and
segments may not capture the full range of artists, musical styles,
vocabulary, lyrical themes, and toxic expressions present in Punjabi music.
Future data collection can therefore expand the number of songs and
segments to improve the coverage and diversity of the resource.

Second, the distribution of the fine-grained toxicity categories is highly
imbalanced. Violence, gang references, and drug references are represented
more frequently, whereas misogyny, casteism, and religious content occur
much less frequently. In particular, only two unique segments were
labelled with casteism, while no unique segment in the final dataset was
assigned to the religious category. This imbalance limits the reliability
of training and evaluating models for the rarest toxicity categories and is
also reflected in the category-specific inter-annotator agreement results.
Future data collection should therefore include additional examples of
under-represented toxicity categories.

Third, the dataset was constructed using publicly available YouTube
sources. Online videos may subsequently be removed, restricted, made
private, or otherwise become unavailable. As a result, access to the
original audio-visual material may change over time. The released resource
therefore provides source identifiers, URLs, and temporal metadata rather
than redistributing the original copyrighted audio and video files.
Researchers who require the original media must obtain it independently
from the relevant lawful sources. Consequently, exact reconstruction of
the original audio-visual inputs may not always be possible if source
content becomes unavailable. Future versions of the resource could
periodically verify source availability and update source metadata where
appropriate.

Fourth, copyright restrictions limit the redistribution of the original
lyric material. The public release does not contain original lyric text or
lyric excerpts. Instead, it provides segment identifiers, temporal
metadata, source information, and annotation labels that allow researchers
to associate the annotations with independently obtained source material.
Researchers must therefore obtain the corresponding lyrics and media
through lawful sources and comply with applicable copyright and platform
terms. This introduces an additional reproducibility constraint because
the availability and content of external sources may change over time.

Fifth, the annotation labels reflect the judgments of four Punjabi-native
speaker annotators and may contain subjectivity, particularly for
context-dependent, metaphorical, or culturally specific expressions. The
shared subset of 158 segments was independently annotated by all four
annotators and was used to quantify inter-annotator agreement using Fleiss'
Kappa. Although the overall toxicity agreement was substantial, agreement
varied considerably across annotation dimensions. In particular, sentiment
agreement was relatively low ($\kappa = 0.2508$), which may reflect the
difficulty of assigning sentiment to short, context-dependent, and
potentially ambiguous song segments. Category-specific disagreement was
also observed for rare toxicity categories. Future work could involve a
larger and more diverse annotation team, additional expert review of
difficult cases, and expanded annotation of under-represented categories.

Finally, the current resource primarily represents Punjabi--English
code-mixed lyrics and Roman-script transcriptions. Consequently, it does
not comprehensively represent all Punjabi linguistic varieties, scripts,
dialects, or regional forms. Future versions of the dataset could expand
coverage to additional Punjabi language varieties, scripts, dialectal forms,
and patterns of code-mixing.

Future work will focus on expanding the dataset, improving the
representation of rare toxicity categories, increasing annotation coverage,
and evaluating the resource across different modelling approaches.
Additional benchmark experiments can be conducted using text, audio, video,
and multimodal approaches to establish reproducible baselines for Punjabi
toxicity detection. Such extensions would allow the resource to support
broader investigation of multimodal toxicity in Punjabi musical content.

\section{Conclusion}
\label{sec:conclusion}

This work presents PGT-FG, a multimodal resource developed to support
research on toxicity detection in Punjabi musical content. The dataset
contains 2,396 unique song segments from 397 songs, with annotations covering
binary toxicity, fine-grained toxicity categories, severity, sentiment, mode,
and annotation uncertainty. The multi-label annotation scheme allows a
segment to represent multiple forms of potentially harmful content, while a
shared subset of 158 segments independently annotated by all four annotators
was used to assess inter-annotator agreement using Fleiss' Kappa.

The dataset provides a resource for research in Punjabi natural language
processing, toxicity detection, and multimodal content analysis. Its
segment-level annotations support the study of potentially harmful content
in musical contexts, where textual, audio, and visual information can
provide complementary signals. The dataset also documents important
limitations, including the imbalance of fine-grained toxicity categories,
the limited representation of rare categories, annotation subjectivity, the
relatively low agreement observed for sentiment annotation, and the changing
availability of online source media.

The annotation resource and associated research metadata are publicly
available through Zenodo under the CC BY 4.0 license at
\url{https://doi.org/10.5281/zenodo.23040156}. The public release contains
2,396 resolved segment-level annotation records from 397 songs, while the
original 2,870 annotator-level annotation records generated during the
annotation process are not included in the public release. Copyrighted
audio, video, and original lyric text are also not redistributed. The
dataset is intended to provide a reproducible resource for future research
on Punjabi toxicity detection and multimodal analysis.

% ============================================================
% Declarations (required by Springer journals, including LRE)
% Place after the Conclusion and before the References.
% The \providecommand line lets this file compile in both the
% plain article class and the Springer Nature template (sn-jnl),
% which already defines \bmhead.
% Items marked [CONFIRM] must be checked before submission.
% ============================================================

\providecommand{\bmhead}[1]{\paragraph*{#1}}

\section*{Declarations}

\bmhead{Funding}
The authors did not receive support from any organization for the submitted work. % [CONFIRM: if any grant, fellowship, or institutional seed funding was used, replace with: "This work was supported by <funder> under grant <number>."]

\bmhead{Competing interests}
The authors have no competing interests to declare that are relevant to the content of this article.

\bmhead{Ethics approval}
% [CONFIRM: keep ONE of the two options below and delete the other.]
% Option A (approval obtained):
% This study was reviewed and approved by the Institutional Ethics Committee of <institution> (reference number <number>).
% Option B (approval not required):
This study did not involve the collection of personal data from human participants. The annotation task was carried out by members of the research team, who were informed in advance about the potentially offensive nature of the material and could pause or withdraw at any time. The source material consists of publicly available music videos, and no copyrighted audio, video, or lyric text is redistributed. Formal ethics approval was therefore not sought.

\bmhead{Consent to participate}
All annotators participated voluntarily and were informed about the nature of the content before beginning the annotation task. No personally identifying information about the annotators is included in the dataset or reported in this article.

\bmhead{Consent for publication}
Not applicable.

\bmhead{Data availability}
The PGT-FG dataset is publicly available on Zenodo at \url{https://doi.org/10.5281/zenodo.23040156} under the Creative Commons Attribution 4.0 International (CC BY 4.0) license. The release contains the resolved segment-level annotations, segment and song metadata, source YouTube URLs with timestamps, and documentation. Copyrighted audio, video, and lyric text are not redistributed and must be obtained by users from the original public sources.

\bmhead{Code availability}
No source code is included in the current release. % [CONFIRM: if the statistics or Fleiss' Kappa scripts are later uploaded, replace with: "The scripts used to compute the dataset statistics and inter-annotator agreement are available at <link>."]

\bmhead{Author contributions}
% [CONFIRM every role below; adjust to reflect what each author actually did.]
\textbf{Lakhjeet Kaur}: Conceptualization, Data curation, Methodology, Formal analysis, Investigation, Software, Visualization, Writing -- original draft.
\textbf{Jaspreet Kaur}: Data curation, Investigation (annotation), Writing -- review and editing.
\textbf{Komal Rani}: Data curation, Investigation (annotation), Writing -- review and editing.
\textbf{Yougal Attri}: Data curation, Investigation (annotation).   
\textbf{Ramakant Kumar}: Conceptualization, Methodology, Validation, Writing -- review and editing.
\textbf{Jaydeep Kishore}: Conceptualization, Investigation, Writing -- review and editing.
\textbf{Pravin Kumar}: Conceptualization, Methodology, Supervision, Project administration, Writing -- review and editing.
All authors read and approved the final manuscript.

\bibliographystyle{unsrt}
\bibliography{references}

\end{document}